# Outlier Detection using AI: A Survey

Md Nazmul Kabir Sikder *and* Feras A. Batarseh

Bradley Department of Electrical and Computer Engineering (ECE), Virginia Tech, USA

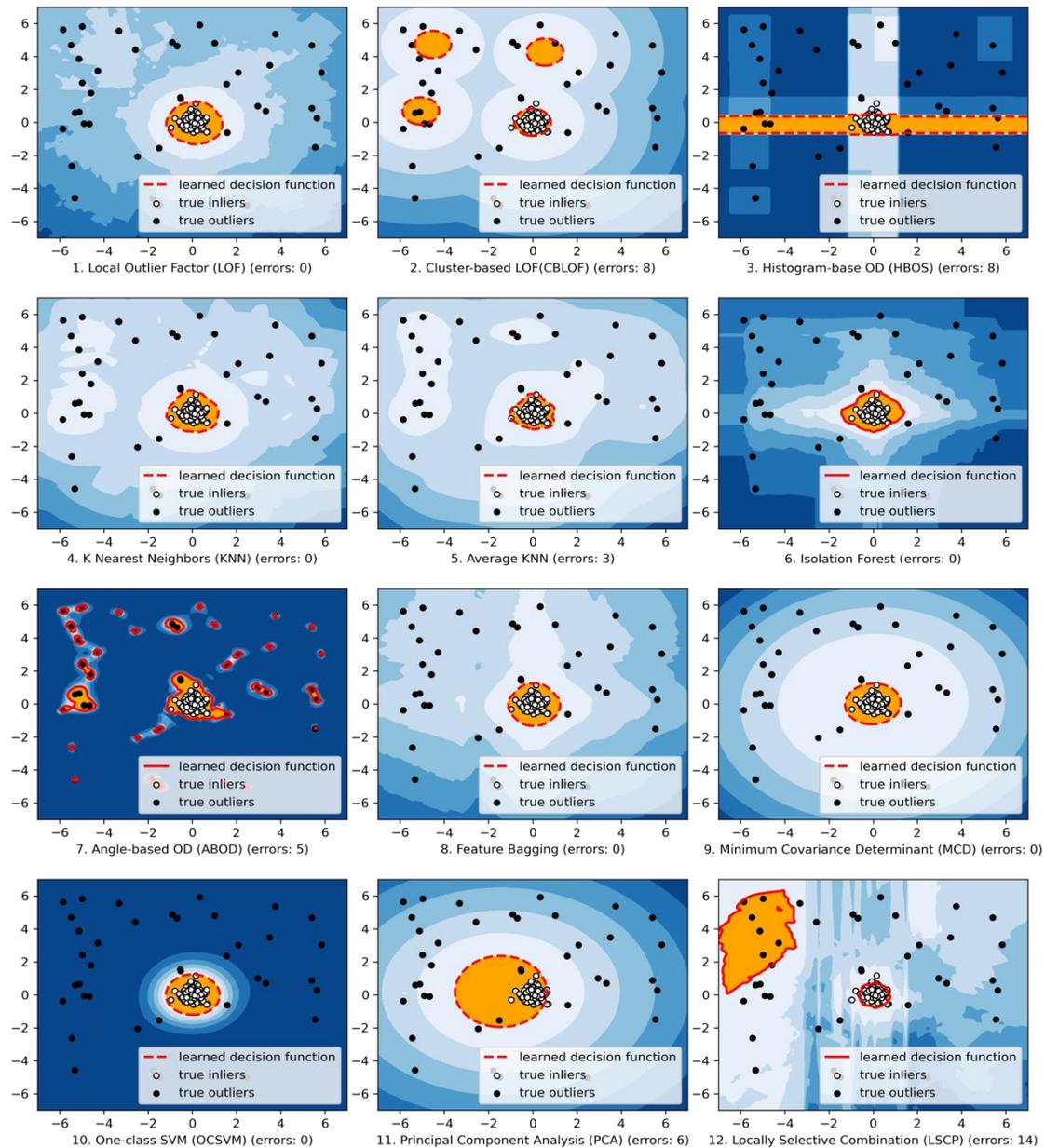

*Data Source: 200 samples created using NumPy library, where outlier contamination is 20%*

# Highlights

**Outlier Detection using AI: A Survey**

Md Nazmul Kabir Sikder, Feras A. Batarseh

- A comprehensive review of outlier detection algorithms from the perspective of Artificial Intelligence (AI)
- Multiple outlier detection categories are introduced and relevant studies are reviewed
- Advantages, disadvantages, research gaps, and suggestions are addressed for each outlier detection category
- AI Assurance is defined and discussed in relation with outlier's detection and analysis

# Outlier Detection using AI: A Survey

Md Nazmul Kabir Sikder
Virginia Tech
[nazmulkabir@vt.edu](nazmulkabir@vt.edu),

Feras A. Batarseh
Virginia Tech
[batarseh@vt.edu](batarseh@vt.edu)

ABSTRACT

An outlier is an event or observation that is defined as an unusual activity, intrusion, or a suspicious data point that lies at an irregular distance from a population. The definition of an outlier event, however, is subjective and depends on the application and the domain (Energy, Health, Wireless Network, etc.). It is important to detect outlier events as carefully as possible to avoid infrastructure failures because anomalous events can cause minor to severe damage to infrastructure. For instance, an attack on a cyber-physical system such as a microgrid may initiate voltage or frequency instability, thereby damaging a smart inverter which involves very expensive repairing. Unusual activities in microgrids can be mechanical faults, behavior changes in the system, human or instrument errors or a malicious attack. Accordingly, and due to its variability, Outlier Detection (OD) is an ever-growing research field. In this chapter, we discuss the progress of OD methods using AI techniques. For that, the fundamental concepts of each OD model are introduced via multiple categories. Broad range of OD methods are categorized into six major categories: Statistical-based, Distance-based, Density-based, Clustering-based, Learning-based, and Ensemble methods. For every category, we discuss recent state-of-the-art approaches, their application areas, and performances. After that, a brief discussion regarding the advantages, disadvantages, and challenges of each technique is provided with recommendations on future research directions. This survey aims to guide the reader to better understand recent progress of OD methods for the assurance of AI.

# 1. Introduction and Motivation

An outlier or anomaly can be defined as abnormality, deviant, or discordant data point from the remaining data set in data science literature. According to (Hawking, 1980, pp. 19), *"an outlier is an observation which deviates so much from the other observations as to arouse suspicions that it was generated by a different mechanism"*. During the development of AI-based applications, data are being created by several generational processes or observations collected from one or multiple entities. Outlier points generate when one or a collection of entities behave in an unusual manner. Therefore, it is very important to understand the behavior of outliers to diagnose a system's health and predict potential system failures. Some of the most popular OD applications are intrusion detection methods (Alrawashdeh and Purdy, 2016), credit card fraud detection (Porwal and Mukund, 2018), medical diagnosis (Gebremeskel and Haile, 2016), sensor events in critical infrastructure, precision agriculture, earth science, and law enforcement (Bordogna et al. 2007). One of the recently succesful example applications of OD is credit card fraud identification, where an AI algorithm is used to find if sensitive information, such as customer identification or a card number is fraudulent or stolen. In such contexts, unusual buying patterns are observed, especially large transactions or irregular buying activities.

In networking and the Internet of Things (IoT) domain, sensors are frequently used to detect environmental and geographical information; changes in underlying patterns, if they occur suddenly, might indicate important events. Event detection in sensor networks is one of the most compelling applications in cyber-physical system. Another OD example is from medical diagnosis, where data are collected from numerous medical devices, including MRI (Magnetic Resonance Imaging) scans, PET (Positron Emission Tomography) scans, and ECG (Electrocardiogram) time-series where unusual patterns could indicate an illness.

In data mining literature, normal data are also known as "*inliners*" (Aggarwal, 2017). Often in real-world applications, such as fraud or intrusion detection system, outliers are *sequential* and not single datapoints within a sequence. For instance, network intrusion is an event in a sequence that is intentionally caused by an individual. Properly identifying the anomalous event helps to handle those sequences.

In most conventional cases, OD algorithms have two types of outcomes: binary labels and outlier scores. Outlier scores impose the level or degree of "*outlierness*" of each data point. Scores naturally rank outlier points and provide various information about the algorithm. However, they don't represent a concise summary with small group sizes. Binary labeling is used to represent if a datapoint is a strong outlier or an inliner. Algorithms can directly provide binary labeling or other means of labeling such as outlier scores, which then can be converted to binary labels for learning purposes. For that, a threshold is selected based on the statistical distribution of the dataset. Binary labels provide less information regarding the degree of outlierness, however in most applications, it is the desired outcomes for decision making process.

For an outlier, defining how much *deviation* is sufficient from a normal datapoint is a subjective judgment. Datasets from real applications might contain embedded noise, and analysts might not be interested in keeping such noise. Therefore, investigating significant deviation is a prime decision to make for OD algorithms. To comprehend this problem clearly, Figure 1(a) and 1(b) illustrate two-dimensional feature spaces. It is evident that clusters are identical in both figures. However, considering a single datapoint "A" in Figure 1(a) seems different from the rest of the datapoints. Therefore, "A" in Figure 1(a) is clearly an outlier. However, point "A" in Figure 1(b) is surrounded by noise and it's quite difficult to say if it is noise or an outlier. When designing algorithms, normal and outlier boundary conditions need to be precise and specific to application requirements.

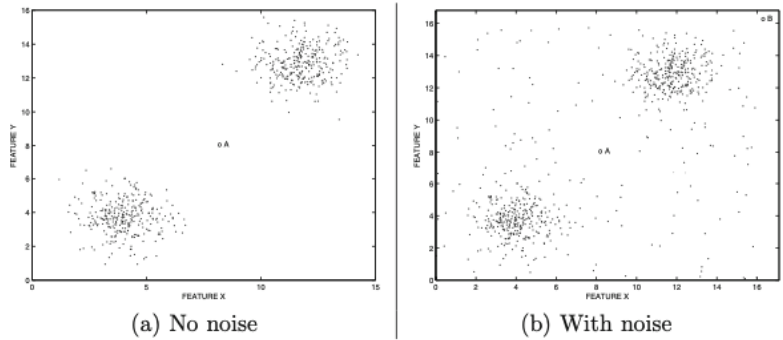

Figure 1: Anomalies and noise in data- Image source: (Aggarwal, 2017)

In unsupervised learning models, noise is defined as weak anomalies that don't hold criteria of being an outlier. For instance, datapoints close to the boundary are mostly considered noise (as presented in Figure 2). Often the separation criteria of these datapoints is subjective and depends on the interest of application-specific demands. Real datapoints that are generated from noisy environments are difficult to detect using scores. That is because noise represents deviated datapoints, therefore requires domain experts to select the threshold between noise and outliers to satisfying application requirements.

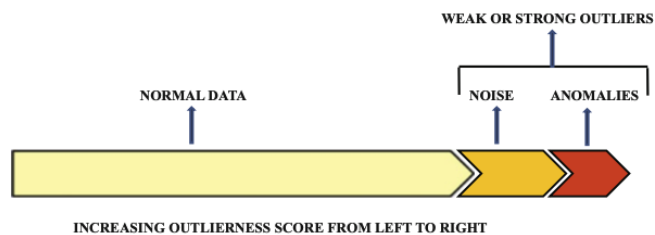

Figure 2: A Typical Data Spectrum with Noise and Outliers

Success in OD depends on data modeling, where every application has its own unique data management requirements. Evidently, the OD technique needs to process the attribution in the data and be sensitive enough to understand the underlying data distribution model. By properly examining the data model, contextual outliers can be achieved. Aggarwal et al. (2011) proposed a concept of linkage outlier by analyzing social networks. Here, nodes that don't show any connection with each other are likely to be outliers, therefore data distribution models play an important role for designing OD models.

OD is a creative process; many researchers are trying to answer the question of how to identify outliers. Research communities are trying to bring forward many innovative and novel algorithms for OD (Aggarwal et al. 2016; Hadi et al. 2009). While identifying and removing outliers from the dataset, researchers need to be very observant, because sometimes outliers carry important hidden information about data. It is crucial to understand data types applying OD methods, for instance, data can be univariate or multivariate and need different approach to begin with. In statistical analysis, careful observation regarding feature selection needs to be considered because we usually want the feature to represent the data distribution model for both non-parametric and parametric analysis. Moreover, during OD, one must make analytic arguments and intuitions before making any conclusions. Besides, real world applications require context-aware and purpose-based detection because the outcome of the result should benefit the requirements of outlier analysis in any given domain. Some recent state-of-the-art application areas are as follows:

**Fraud and Intrusion Detection:** Intrusion detection is performed to check if a computer network has any unauthorized access by observing unusual patterns (Singh et al. 2010). Additionally, to make a network secure and safe, detection of outlier instances is extremely important.

**Database and Sensor Network Monitoring:** Sensor networks require continuous monitoring for effective wireless operations. Detecting outliers in sensor network (Abid et al. 2017; Feng et al. 2017), body sensor networks (Zhang et al. 2016), and target tracking environments (Shahid et al. 2015) ensures flawless operations with proper routing in the network.

**IoT and Critical Infrastructure Operations:** IoT devices utilize wireless sensors to collect various information on architecture including smart grid, power distribution system, water supply system, and healthcare diagnostic system. It's very crucial to know correct and effective data are being collected form IoT devices. If the data are being polluted with outliers because of a sensor fault or a cyber-attack, that should be identified for securing the critical infrastructure. Additionally, OD algorithms need to be trained against attack concealment. Critical Infrastructures are the backbone of society; effective and efficient OD models are crucial for optimal operations, preventive maintenance, and the overall safety and security of our nation.

**Data Streams Monitoring:** Zheng et al. (2016); Tamboli and Shukla (2016); Shukla et al. (2015); Tran et al. (2016) and Gupta and Gao (2014); Cateni et al. (2008) showed OD for data streams and time series datasets. Detecting outliers in data streams is important because any abnormality may hinder fast computational and estimation processes of applications.

**Medical Diagnosis:** Modern healthcare and diagnosis analysis are mostly dependent on electronic devices. These devises observe unusual patterns while reading different measures from patients. Properly separating anomalous readings help doctors to predict underlying conditions and thereby to apply proper diagnosis.

**Fake News Detection:** Fake news can be considered as an outlier if compared with foundational datasets and real sources (Shu et al. 2017).

**Surveillance and Security:** Security is an important aspect in computer administrative network. Cybersecurity is a field where researchers ensure methods for safe access and proper authentication. An exciting and practical research in cybersecurity is surveillance video OD (Xiao et al. 2015).

**Data Logging and Data Quality:** Logging and processing data for commercial purposes can go wrong because of unwanted concealment processes, which if not detected, might result in irrecoverable loss. Automated data mining models are applied in searching for abnormalities while processing large volume of logs (Ghanbari et al. 2014). Proper anomaly identification algorithms need to be applied to enhance data quality (D'Urso et al. 2016; Chenaoua et al. 2014).

The rest of the chapter is organized as follows: In section 2, we categorize OD algorithms into six subgroups, where each subgroup has a detailed discussion, advantages, disadvantages, research gaps and suggestions. In section 3, we include multiple OD tools. In section 4, we enlist several benchmarking datasets for outlier analysis and in section 5, we discuss AI assurance and its relevance to outlier analysis. Finally, in section 6, we conclude with open research gaps and OD challenges.

## 2. Outlier Detection Methods

OD methods can be classified into many categories (Ranshous et al. 2015; Braei and Wagner, 2020; Lai et al. 2021), however, in this chapter we introduce six major categories: Statistical, Density, Clustering, Distance, Learning, and Ensemble-based OD methods. For each group, we provide short overview about their gradual development over the last few decades.

### 2.1. Statistical and Probabilistic Based Methods

Statistical and probabilistic based OD methods originated from early nineteenth century (Edgeworth et al.1887). Before inventing high performance devices these methods were applied for simple data visualization, although performance and efficiency were being neglected. Nevertheless, the fundamental mathematics are always useful and eventually these methods are applied to most regular OD applications.

Almost all the OD algorithms apply numerical scores to every object and in the final step they assign extreme values by observing the scores. Binary classification is one way of sorting the extreme value points. Statistical and probabilistic OD algorithms can be either supervised, unsupervised or semi-supervised. The model is built based on data distribution. For statistical-based OD algorithms, stochastic distribution is a widely adopted technique to detect outliers. Therefore, the degree of outlierness depends on the model built using data distribution. Statistical and probabilistic based methods can be further divided into two broad categories: parametric and non-parametric distribution models. Parametric methods assume a distribution model from the dataset and then use knowledge from the data to approximate model parameters. Non-parametric methods don't assume any underlying distribution model (Eskin et al. 2000).

### 2.1.1. Parametric Distribution Models

Parametric distribution models have prior knowledge of the data distribution, these models can be divided into two subcategories: Gaussian Mixer and Regression models.

**Gaussian Mixture Models:** Gaussian model is a popular statistical approach in OD, it initially adopts Maximum Likelihood Estimation (MLE) in training stage to compute variance and mean of the Gaussian distribution. During the test phase, several statistical measures are applied (mean variance test, box plot test) to validate the outcomes.

Yang et al. (2009) proposed an unsupervised Gaussian Mixture Model (GMM) based on an explainer that globally optimizes to detect outliers. In this method, first it fit the GMM for a dataset by utilizing the Expectation Maximization (EM) algorithm based on global optima. Outlier factor for this method is calculated as the sum of proportional weighted mixture, the weights represent affiliations to remaining data points. Mathematically, outlier factor can be expressed at $x_k$:

$$F_k = z_k(t_h) = \sum_{j=1}^{n} s_{kj} \pi_j(t_h) \qquad (1)$$

where, $s_{kj} \pi_j(t_h)$ = Point $X_k's$ relationship with other point $X_j$.
$s_{kj}$ = Relationship strength
$t_h$ = Iteration (Final)
$\pi_j$ = Degree of importance of point $j$

Higher outlier factor indicates greater degree of outlierness. This method focuses on global properties rather local ones that we discuss later in density-based method section (Breunig et al. 2000; Papadimitriou et al. 2003; Tang et al. 2002). Yang et al. (2009) claimed, for a given dataset, fitting the GMM at each data point- outlier can be detected even if the dataset contains noise, which was a major challenge in clustering-based techniques. Therefore, this technique is useful in real-world applications, where environmental noise or intentional adversarial noise is embedded. It is evident that the algorithm has higher capacity to detect unusual objects, however it incurs greater complexity: for single iteration model complexity is $O(n^3)$ and for $N$ iteration model complexity is $O(Nn^3)$. Future studies shall improve the algorithm and reduce its computational complexity along with increasing its scalability.

Tang et al. (2015) proposed an improved and robust statistical model, they applied GMM with projections preserving locally. They applied the model to disaggregate energy utilization by combining both outcome of Subspace Learning (SL) and GMM. In this method, the LPP short for locality preserving projection of SL is exploited to reveal the inherent diverse structure while at the same time keeping the neighborhood composition intact. Saha et al. (2009) proposed a Principality Component Analysis (PCA) technique that points research gaps in Local Outlier Factor (Breunig et al. 2000) and Connective-based Outlier Factor (Tang et al. 2002) that fails to achieve multi-Gaussian and multiple state OD. The method shows improved performance, however, the authors barely discussed anything about their model's computational complexity.

**Regression Models:** Regression OD models, depending on the context, are either linear or non-linear. They are a direct approach to detect outliers. Generally, the training stage involves fitting the given datapoints into a constructed regression model. The regression models are evaluated at the test stage. Outliers are labeled if the difference between actual output and predicted outcome of the regression model is too high. For last few years, OD using regression analysis applied several standard techniques as Mahalanobis distance, mixture models, robust least squares, and Bayesian alternate vibrational methods (Zhang, 2013). Satman (2013) in contrast to other algorithms, proposed a different one, on that has a covariant matrix which is non-interactive. It has less computational complexity, which makes it cost effective as it can detect multiple outliers quickly. For future research directions, and as regression models often portrayed as minuet preference, variance and bias of the intercept approximator can be minimized to improve the result.

Another regression model proposed by Park and Jeon (2015) detects outliers in sensor network. The method observes the values from the model outcome and create an independent variable using a weighted sum approach. Since the model only applied on a single sensor environment, measuring outliers accurately from multiple sensor environment can be an interesting topic (as a future direction). Dalatu et al. (2017) studied a comparison between linear and non-linear model, where their accuracy and misclassification were examined with Receiver Operating Characteristic (ROC) curves. This case study provided necessary information for OD for two popular kinds of regression models. Non-linear models showed more accuracy (accuracy 93%) compared to linear regression models (accuracy 63%), therefore it's mostly a better option to select non-linear models over linear regression models.

### 2.1.2. Non-Parametric Distribution Models

Non-parametric distribution models don't assume any underlying data distribution (Eskin, 2000) for given datasets. Kernel Density Estimation (KDE) models are a popular non-parametric approach, they are unsupervised technique to detect outliers that utilizes kernel functions (Latecki et al., 2007). The KDE model compares each objects density with neighbors' densities, where the idea is similar as some of the prevalent density-based techniques (Papadimitriou et al. 2003; Breunig et al. 2000). Although, it has improved performance, the curse of dimensionality reduces its applicability. Gao and Hu (2011) offered a superior solution to overcome the problem. They applied kernel-based technique that has lower run time compared to (Latecki et al. 2007, Breunig et al. 2000), also presented better scalability and performance for large data sets. This method solves another limitation of Local Outlier Factor (Breunig et al. 2000): sensitivity on parameter k, where it measures the weights of local neighborhoods by utilizing weighted neighborhood density estimations.

A good real-world application by Samparthi and Verma (2010), also applied KDE to measure infected nodes in a sensor network. Boedihardjo et al. (2013), in another study implement the KDE method in time series dataset, although it was a challenge using KDE for data streams. They proposed an accurate estimation of Probability Density Function (PDF) by using adaptive KDE. The computational cost associated with the method is $O(n^2)$, and showed better estimation results compared to original KDE. The method is suitable for strict environment, therefore further research may improve the method for adopting multivariate data. Uddin (2015) applied the KDE method in power grid environment. Although, the KDE methods are better at targeting outliers, they are computationally expensive. Later, Zheng et al. (2016) applied KDE in a multimedia network for outlier detection on multivariate dataset. In another study, Smrithy and Munirathinam (2016) introduced a non-parametric method for outlier detection in big data. Another nonlinear system is studied by Zhang et al. (2018), proposed a technique based on Gaussian Kernel called adaptive kernel density-based approach. Later, Qin and Cao (2019) proposed a unique OD approach that perfectly applies KDE to effectively identify local outliers from continuous datasets. This method facilitates to detect outliers from high data stream irrespective of data complexity and unpredictable data update, which was a challenge earlier. They derived an approach to successfully identify top-N outliers based on KDE on continuous data. Afterwards, Ting et al. (2020) modified the KDE approach to identify similarity between two distribution named Isolation Distribution Kernel. Compared to other Kernel based algorithm, the proposed method outperforms most point anomaly detection. Although, KDE based approach performs better compared to other non-parametric models, they suffer from high dimensionality in the feature space. Additionally, in general they have high computational cost too.

### 2.1.3. Miscellaneous Statistical Models

Among many proposed OD algorithms, most straightforward techniques in statistical method are Trimmed Mean, Boxplot, Dixon test, Histogram, and Extreme Studentized Deviate (ESD) test (Goldstein and Dengel, 2012; Walfish, 2006). The Dixon test works well with small size dataset as no assumption is required about data normalcy. The Trimmed Mean is not a good approach among all others for OD, however ESD test is a better choice. Pincus (1995) introduced several optimization tests for OD that could depend on parameters such as number and expected space of outliers. A histogram-based OD technique-HBOS is proposed by Goldstein and Dengel (2012), which can create model of univariate feature space by utilizing dynamic and static histogram bin width. Here, each data point is scored as degree of outlierness. The algorithm showed improved performance, especially faster computational speed over traditional OD approaches as (Jin et al. 2006; Tang et al. 2002; Breunig et al. 2000). Nevertheless, the method faces difficulties finding local outliers with its density approximation technique. Hido et al. (2011) introduced a novel statistical methodology by applying guided density ratio approximation to detect outliers. The main idea of the algorithm is to select density ratio between training set and test set. A natural cross validation method was applied to optimize the value of parameters: regularization and kernel width. To achieve better cross validation performance, unconstrained least square method was applied. This method has an advantage over non-parametric kernel density estimation because hard density estimation isn't required here. The method in terms of accuracy, shows improved performance in most cases. Improving density ratio estimation of this method is an important research direction.

Robust Local Outlier Detection (RLOD), another method that adopts statistical measures to detect outliers is proposed by Du et al. (2015). This pipeline assumes the fact that OD is sensitive to parameter tunning (Gebhardt, and Goldstein, 2013) and most OD methods are focused to detect global outliers. The whole pipeline can be divided into three stages. At first stage, it applies three standard deviation measures to find density peaks of the dataset. In $2^{nd}$ stage, remaining data points are labeled to the closest higher density neighbors by assigning them in matching clusters. In the $3^{rd}$ and final stage, it applies density reachability and Chebyshev's inequality to locate local outliers for each collection. Campello (2015) showed that RLOD can both detect local and global outliers, they experimentally showed that RLOD outperforms some former OD algorithms (Breunig et al. 2000; Zhang, 2013) in terms of detection rate and execution time. RLOD performance can be improved more by adopting parallel and distributed computing. Later in another study, Li et al. (2020) proposed an effective Copula-Based OD.

### 2.1.4. Advantages of Statistical and Probabilistic based Methods

The fundamental mathematics behind statistical OD algorithms make them easy to use. Due to their compact form, the models exhibit improved performance in terms of detection rates and run times for a particular probabilistic technique. For quantitative ordinal and real-valued data distribution, the models usually fit well, although results could be more improved if ordinal data can be preprocessed. Despite some targeted issues such as high dimensional feature space, the models are convenient to deploy.

### 2.1.5. Disadvantages of Statistical and Probabilistic based Methods

The parametric models assume underlying density distribution, which results in poor performance and often might bring unreliable outcomes in real-world applications such as managing data streams from a complex network. Statistical-based approach is applicable mostly for univariate datasets; therefore, they don't perform well for multivariate feature spaces. If the models are applied to multivariate feature space, high computational cost incurs, which make them a poor choice for multivariate data stream. Additionally, the histogram cannot capture the interaction between features which makes it a poor choice for high dimensional data as well. Therefore, statistical methods that can investigate simultaneous feature space can be promising research. To deal with the curse of high dimensionality, specific statistical methods can be adopted, however it results in longer processing time and a misleading data distribution.

### 2.1.6. Research Gaps and Suggestions

Several common research gaps in statistical-based approach are poor accuracy, difficulties with high dimensional datasets, and operational expense. These gaps need to be addressed in the future to make the models more reliable. These methods however can be more effective if applied model is aware of the context. Time series data generated from critical infrastructures such as smart grid and water distribution system may contain anomalous samples because of maintenance problems or intentional attacks, however their pattern is unknown to a learning model. In this scenario, parametric methods fail to learn the underlying distribution as it constructs the model based on predefined data distribution. Therefore, for this case non-parametric methods are a better choice, as they don't need to know the underlying distribution of a given dataset. Also, parametric methods are not a better choice for large data stream where outlier points are dispersed evenly. Inaccurate labeling of outliers might occur if the threshold is defined based on standard deviation to separate them. Using parametric methods for OD is a difficult task while applying GMM to manage data stream and high dimensional feature space. Therefore, algorithms that can easily manage data stream along with high dimensional feature space can make the model more scalable. High dimensionality also creates problem for regression models. To overcome this issue, targeted regression analysis can be adopted instead of ordinary regression analysis.

Non-parametric models, especially KDE are a better choice in most applications, however they get computationally expensive in noisy environments. In contrast with parametric methods, KDE is scalable, although computationally expensive for multivariate data. The histogram-based approach is a good fit for univariate data distribution, however its inability to investigate the interaction among features makes it a poor choice for multivariate data. Despite statistical methods inability to adopt some recent application areas, they are still a good choice for targeted domain and data streams. PCA methods by Saha et al. (2009) and Tang et al. (2015) are effective approaches for OD. Goldstein and Dengel (2012) proposed a Histogram-Based Outlier (HBOS), it shows improved performance when compared to other clustering-based models such as Local Outlier Factor, Local Correlation Integral, and Influenced Outlier in terms of calculation speed, therefore is a good choice for real time data (Breunig et al. 2000; Papadimitriou et al. 2003; Jin et al. 2006); OD models scalable to large dataset proposed by Du et al. (2015) and Hido et al. (2011) also proved robust in analyzing outliers.

### 2.2. Density-Based Methods

Density-based OD is one of the most popular and prevalent techniques. The main principal is that an outlier point can be found in a sparse region whereas normal points can be found in denser region. Figure 3 presents a two-dimensional dataset where labeled point "A" and "B" are considerably separated from the rest of the densely populated clusters, therefore are outlier points in this dataset. The core idea for detecting outlier points "A" and "B" is that these points remain in sparse populations, whereas the normal points are in higher denser populations. Density-based methods seek for differences between densities of a point with their local neighborhood. Usually, density-based methods are computationally expensive compared to distance-based methods. Despite this problem, density-based methods are widely popular because of their simplicity and efficiency to detect outliers. Some baseline algorithms utilizing these methods are presented in Breunig et al. (2000); Jin et al. (2006). Zhang et al. (2009); Tang and He (2017) presented algorithms that are developed and modified version of those baseline one's.

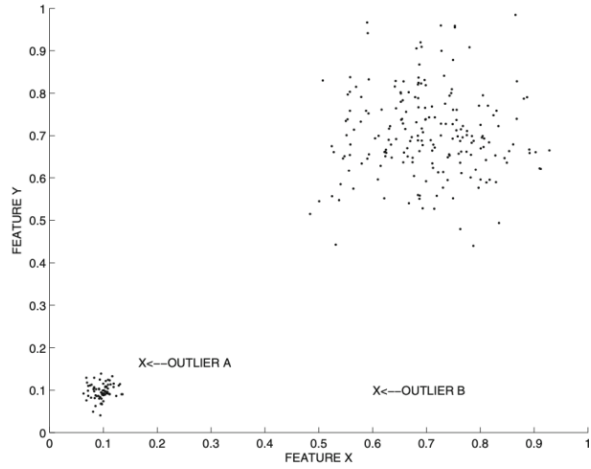

Figure 3: Density-based outlier detection (Aggarwal, 2017)

**Local Outlier Factor (LOF):** LOF is a popular method proposed by Breunig et al. (2000), which is the base algorithm that represents density-based clustering method for detecting outliers. K-Nearest Neighbor (KNN) technique is used in this process, for each point in a KNN set, LOF measures local reachability density (lrd) to differentiate each point with its neighborhood. Mathematically, $lrd$ can be defined as:

$$lrd(p) = \frac{1}{\frac{\sum_{o \in kNN(p)} reach-dist_k(p \leftarrow o)}{|kNN(p)|}} \quad (2)$$

$$\text{LOF score: } LOF_k(p) = \frac{1}{|kNN(p)|} \sum_{o \in kNN(p)} \frac{lrd_k(o)}{lrd_k(p)} \quad (3)$$

where, $lrd_k(p)$ = $lrd$ of point p
$lrd_k(o)$ = $lrd$ of point o

The main idea of the LOF is to determine the degree of outlierness of an observation while comparing its cluster with local neighbors. The LOF score gets higher for an observation if it's $lrd$ value is less than the estimated nearest neighbor. Logging $lrd$ value and computing LOF score using KNN approach costs $O(k)$ for each data point. It is wise to use a valid index because a sequential search of a size $n$ dataset can cost $n^2$ if a proper indexing is not applied.

Schubert and Zimek (2014) addressed this shortcoming and introduced a simplifiedLOF method which makes the density estimation simpler. The simplifiedLOF method adopts KNN distance instead of LOF's reachability distance.

$$dens(p) = \frac{1}{k-dist(p)} \quad (4)$$

The simplifiedLOF is more computationally complex than LOF but improved in performance.

**Connective-based Outlier Factor (COF):** Tang et al. (2002) realized an improved method, COF over methods proposed by Breunig et al. (2000); Schubert and Zimek (2014). The COF is almost similar as the LOF, although density estimation calculation is different. The COF applies chain distance to calculate local densities of neighbors, but Euclidian distance is generally applied to LOF. Because of applying chaining distance for density estimation, this process assumes predefined population distribution, which is a major drawback, because it often results in wrong density estimation. The authors applied a new term- "isolativity" instead of "low-density" to locate outliers. Isolativity is a unique measure that represents the degree of connectedness of an observation with the remaining points. At point p, the COF value can be expressed mathematically while applying the KNN approach is:

$$\text{COF}_k(p) = \frac{|N_{k(p)}|ac - dist_{Nk(p)}(p)}{\sum_{o \in Nk(p)} ac - dist_{Nk(p)}(p)} \qquad (5)$$

where $ac - dist_{Nk(p)}(p) =$ Average chain distance between point p and $N_{k(p)}$

In the neighborhood, COF modifies density estimation of the SimplifiedLOF to verify the connectedness using a method called Minimum Spanning Tree (MST). The computational cost is $O(k^2)$ that occurs for calculating MST from KNN set. Except in circumstances, where data sets are characterized by connective data patterns, COF takes similar time as LOF for detecting outliers.

**Local Outlier Probabilities (LoOP):** The LOF algorithm uses scores for each data points of KNN, however threshold selection for labeling datapoints was a growing question. Therefore, Kriegel et al. (2009) proposed LoOP that generates score with statistical probabilistic approach. In this method, density is estimated using distance distribution. LOF scores are presented as statistical probabilities. They compare the advantages of assigning probabilities of a datapoint over outlier score in LOF. Mathematically LoOP can be expressed as:

$$LoOP_s(O) = max\left\{0, erf\left(\frac{PLOF_{\lambda,S}(O)}{nPLOF.\sqrt{2}}\right)\right\} \qquad (6)$$

where, $PLOF_{\lambda,S}(O) =$ LOF probability wrt importance of $\lambda, r$
$nPLOF =$ aggregated value

Normal points that are in denser population will have LoOP value almost zero, whereas LoOP value towards 1 indicated loosely connected points or outliers in the dataset. As simplifiedLOF (Schubert and Zimek, 2014) the LoOP also has same computational complexity for each point: $O(k)$. A significant difference for calculating local densities compared to previous density-based methods is that it assumes and applies half-Gaussian distribution for density estimations.

**Local Correlation Integral called (LOCI):** Papadimitriou et al. (2003) proposed a method called LOCI that correctly handles multi-granularity issue, where LOF (Breunig et al. 2000) and COF (Tang et al. 2002) both were unable to solve the problem. They defined an outlier metric-MDEF short for Multi Granularity Deviation Factor, according to the method outliers are points that is away from the neighbor of MDEF by at least three standard deviations. Not only this method finds both remote cluster and concealed outliers but also deals with feature space local density variation. The MDEF can be defined mathematically on a point $p_i$ within a radius $r$:

$$MDEF(p_i, r, \alpha) = 1 - \frac{n(p_i, \alpha r)}{\hat{n}(p_i, r, \alpha)} \qquad (7)$$

where, $n(p_i, \alpha r) = \alpha r$ neighborhood objects number

$\hat{n}(p_i, r, \alpha) =$ All the objects $p's$ average at $r$-neighborhood of $p_i$

To get faster result from MDEF, right side fraction needs to be measured after getting the value of numerator and denominator. So far, all the OD algorithm we have discussed based on KNN algorithm, detection of number of k is crucial to find outliers properly. The LOCI algorithm is better because it applies a maximization process to find out optimal k-value. It is because for estimating local densities LOCI applies half Gaussian distribution that mimics LoOP (Kriegel et al. 2009). Instead of measuring distances for density estimation, they aggregate the local neighborhood records. Also, LoOP is different because it of its unique neighbor comparison. Although, LOCI shows good results, it has longer run time. Papadimitriou et al. (2003) developed a different approach to increase the speed by introducing quad tree with several constraints between two neighbors.

**Relative Density Factor (RDF):** Ren et al. (2004) proposed a new technique for effective OD by pruning data points located in deep cluster. This algorithm takes advantage of large datasets and provides scalability. RDF adopts a data model to identify anomalies, called *P-tree*. Higher RDF values indicate greater outlier behavior of data points in the population. RDF can be mathematically expressed on point $p$ and radius $r$ as:

$$RDF(p,r) = \frac{DF_{nbr}(P,r)}{DF(P,r)} \qquad (8)$$

where, $DF_{nbr}(P,r) \; and \; DF(P,r)$ are both density factor

**Influenced Outlier (INFLO):** INFLO is another technique based on LOF (Breunig et al. 2000) and proposed by Jin et al. (2006). The method detects outliers by assuming symmetric relationship between neighbors. One shortcoming of LOF (Breunig et al. 2000) is that it fails to correctly define scores for datapoints at cluster border, where the clusters are related closely. INFLO solves this problem by distinguishing different neighborhood of context and reference set. The scores are calculated by both reverse nearest neighbor and KNN. INFLO adopts both reverse nearest neighbors and nearest neighbors techniques to achieve accurate neighborhood distribution. Here outliers are observations that have higher INFLO values.

**High contrast subspace (HiCS):** Almost all the previous algorithms described (LOF, COF, LOCI, and INFLO) suffer when calculating distances of large dimensional feature spaces, however, a method proposed by Keller et al. (2012) for large dimensional dataset can successfully sort and rank outliers and their score: High Contrast Subspace method (HiCS).

**Global-Local Outlier Score from Hierarchies (GLOSH):** Campello (2015) proposed a method that includes beyond local outliers and extends the investigation to detect global outliers. This method applies statistical interpretation to find both local and global outliers. Although, GLOSH isn't a generic algorithm, however often it provides better results. The baseline algorithm is KNN, therefore it has some common setbacks, which can be solved by further improving density estimation.

**Dynamic-Window Outlier Factor (DWOF):** Momtaz et al. (2013) proposed a unique algorithm that detects top $n$ number of outliers by assigning outlier score called DWOF. This method deviates from its ancestor algorithms in density-based methods, however, it closely complements Fan et al. (2009) that proposed the the Resolution-based Outlier Factor (ROF). ROF performs better in terms of accuracy and sensitivity to hyperparameters.

**Algorithms for High Dimensional Data:** With the increment in data volume and complex networks, its highly required to design sophisticated and efficient algorithms. Keeping that in mind, Wu et al. (2014) implemented an algorithm that can handle high dimensional data. The algorithm introduces new technique called RS-forest that is faster and more accurate. It includes one class semi-supervised Machine Learning (ML) model. Later, Bai et al. (2016) proposed a similar technique as Wu et al. (2014), which can discover outliers in parallel. LOF (Breunig et al. 2000) is the base algorithm but a new computing method is introduced called distributed computing for density estimation. This algorithm works in two steps, at first it partitions using grid-based technique and then distributed computing identifies the outliers. Unfortunately, this algorithm doesn't scale well; earlier Lozano and Acuna, (2005) fixed this issue by suggesting a technique called PLOFA (Parallel LOF Algorithm), which improves scalability for big data.

**Other Density-Based Algorithms:** Tang and He (2017) proposed a method to estimate density using kernel density estimation for measuring local anomalies; a scoring process is introduced called Relative Density-Based Outlier Score. The model applies KDE that pays more attention on shared neighbor and reverse neighbors rather than KNN to compute density distribution. Distance measure is same as UDLO (Cao et al. 2014) which is Euclidian distance. However, they need to compare different distance measures to observe the changes before applying this method to real applications. Vázquez et al. (2018) introduced a detection algorithm for data that has low density population, called Sparse Data Observation (SDO). The SDO is a hungry learning algorithm that tries to learn quickly and reduces computation time for each object compared to previous algorithm in density-based methods that we have discussed so far. Ning et al. (2018) proposed relative density-based OD method which is similar method as Tang and He, (2017), it's a new technique to compute neighborhood density distribution. Su et al. (2019) implemented local OD algorithm on scattered dataset, instead of using the term LOF they used Local Deviation Coefficient (LDC), because the LDC focuses on distribution of object and neighbors. The algorithm removes normal points in a safe way and keeps the outlier points as reminder; the process is called RCMLQ short for Rough Clustering based on Lulti-Level Queries. Since, it is pruning the normal objects, it is useful for local OD in large dataset. It showed better efficiency and accuracy over previous local OD algorithms.

### 2.2.1. Advantages of Density-Based Methods

Density-based OD algorithms apply non-parametric method to measure density, therefore they don't assume any predefined distribution model to manage the dataset. LOF (Breunig et al. 2000), LoOP (Kriegel et al. 2009), INFLO (Jin et al. 2006) and DWOF (Papadimitriou et al. 2003) are some of the baseline algorithms that serve as the fundamental model. Density-based algorithms can both identify local and global outliers, which make them useful for real-world application and often outperform other statistical-based algorithms (Wang et al. 1997; Akoglu et al. 2014; Hido et al. 2011). Additionally, the fundamental concept is to estimate neighborhood density that provides more flexibility to investigate crucial outliers which can be easily measured by several other modern OD algorithms. Density-based algorithms also facilitate to exclude outliers from nearby denser neighbors. They hardly require any primary knowledge such as probability distribution which makes the algorithm easy for hyperparameter tunning. In fact, only single hyperparameter tuning brings good results. The algorithms are also useful and efficient when it comes to detect local outliers (Su et al. 2019).

### 2.2.2. Disadvantages of Density-Based Methods

Although some of the density-based algorithm showed good performance, however they are computationally expensive and complicated when compared to many statistical-based methods including ones presented by Kriegel et al. (2009). Also, these methods are sensitive to the shape of the neighbors, when cautiously tuning the size hyperparameter they become computationally expensive including increased runtime. It is also evident from the applications that neighbors varying density creates complicated models and generally generates poor result. Few Density-Based methods such as MDEF and INFLO, because of their complex density estimation process, cannot handle datasets resourcefully such as defining outleirness of an object. Also, density-based models face challenge when it comes to manage high dimensional time series data, however recent algorithms seem to overcome the problems by introducing pruning (Ren et al. 2004) and elimination (Su et al. 2019) techniques – among others.

### 2.2.3. Research Gaps and Suggestions

In general, since density-based OD's are non-parametric methods, sample size is considered small for high dimensional feature space. This challenge can be resolved by resampling the objects to enhance the process. As density-based algorithms are based on k-nearest neighbors, therefore proper selection of hyperparameter $k$ is important to evaluate these algorithms. Generally, computational expense using KNN is $O(n^2)$, however LOCI has greater complexity because of adding an extension - radius r, therefore computational cost becomes $O(n^3)$. So, LOCI, when applied to big data, gets very sluggish to compute OD. Goldstein and Uchida, (2016) compared between LOF and COF. They concluded that applying spherical density estimation using LOF creates a poor-quality process for OD. However, COF applies connectivity feature to estimate density pattern to solve the issue. INFLO, when applying to closely related clusters with varying densities, performs better by generating enhanced outlier scores.

### 2.3. Clustering-Based Methods

Clustering-based OD differentiates between clusters and outlier points. A simple description would be: *each datapoint in given dataset that belongs to a cluster is either an outlier or a normal point*. The goal for clustering is to separate the points from denser and sparse population, generally a sparse region contains most of the outliers. Therefore, most clustering algorithms get outliers as a side product of their analysis. While detecting outliers using clustering-based approach, a score is provided that represents the degree of outlierness of a sample. Outlier score can be calculated using the distance between a data point and nearest cluster centroid. Because of different cluster shape, Mahalanobis is a good distance measure that scale well for the clusters. Mathematically, Mahalanobis distance from datapoint X to cluster distribution with centroid μ and covariance matrix Σ is:

$$MB(X, \mu, \Sigma)^2 = (X - \mu)\Sigma^{-1}(X - \mu)^T \qquad (9)$$

*Here, X = dataset,*
*Σ = covariance matrix,*
*μ = attribute wise means of d dimensional row vector,*

After scoring each datapoint with the Mahalanobis distance, binary labels can be assigned by selecting extreme comparison. Mahalanobis distance can be visualized as the Euclidian distance between a sample and a cluster centroid. This distance measure indicates data locality characteristics by providing statistical normalization.

Figure 4 illustrates the effects of identifying outliers while considering data locality. Here, Euclidean distance measure will consider point "A" an outlier over point "B" because of the normal distance measure. However, Mahalanobis distance, considering data locality provides point "B" as more anomalous than point "A", which makes sense visually (Figure 4). Therefore, defining proper number of clusters and a suitable distance measure results in successful outcome of OD algorithm.

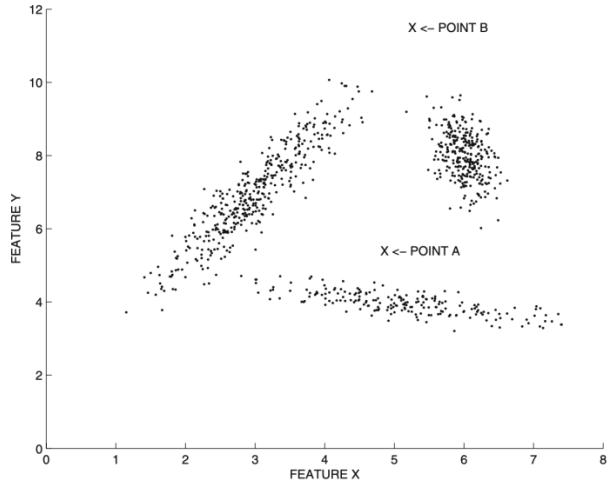

Figure 4: Clustering based OD method (Aggarwal, 2017)

Detecting outlier using clustering-based approach is dependent on properly defining cluster structure of normal instance (Zoubi, 2009), which comes from the effectiveness of the algorithm. These algorithms are *unsupervised* since they don't need any previous knowledge of feature distribution. Many OD techniques are introduces based on clustering algorithms, Zhang (2013) categorized several of them. Clustering based approach is a broad category and can be grouped into several subgroups as following:

**Clustering methods based on partitioning:** These clustering methods are based on distance-based technique where cluster numbers are selected initially or provided randomly. Algorithms belong to this subgroup are presented by MacQueen (1967); Ng and Han (1994); Kaufman and Rousseeuw (2009).

**Clustering methods based on density:** In contrast to partitioning based clustering approach, defining initial number of clusters for these models isn't required. However, they can model the cluster into denser and non-denser groups given the radius of a cluster. Algorithms belong to this subgroup are studied by Hinneburg and Keim (1998) including Density-based spatial clustering of applications with noise (DBSCAN) by Ester et al. (1996).

**Clustering methods based on hierarchy:** In this subgroup, the algorithms partition the cluster into different levels structured like a tree. Algorithms belong to this subgroup are presented by Karypis and Han (1999); Guha et al. (2001); Zahn (1971).

**Clustering methods based on grids:** Algorithms belong to this subgroup are presented by Zhang et al. (2005); Sheikholeslami et al. (2000); Wang et al. (1997).

**Clustering methods based on high dimensional features:** Algorithms belong to this subgroup are presented by Aggarwal (1998), (2004). Besides that, Cao et al. (2006) proposed a two-stage algorithm called DenStream. They applied density-based approach for both offline and online OD. First stage summarizes the given time series dataset then second phase organizes clusters from the summarized data. The DenStream creates a microculture to separate outliers and normal data points. A micro cluster is a real outlier if its weight is less than the predefined threshold and being pruned by the model afterwards. The authors performed a comparison between DenStream and CluStream (Aggarwal and Han, 2003) to present their models effectiveness, where DenStream shows improved performance because it avoids using memory space and utilizes taking snapshots on a disk. However, the model face difficulties when adjusting dynamic parameters in time series datasets and locating arbitrary cluster shapes with multiple levels of granularity. Solving these issues can be a good future study. Later, Chen and Tu (2007) proposed an algorithm like the DenStream regarding offline and online OD, called D-Stream; the only difference is that D-Stream is a grid-based OD algorithm. Outliers, compared to previous algorithm, can be found easily by exploiting the definition of noise in terms of dense, sparse, and sporadic grid. A density threshold is selected to which the sporadic grids are compared, if less than the threshold the datapoints are considered outliers. Also, the algorithm performs better in terms of clustering and runtime compared to CluStream. In another study, Assent et al. (2012) implemented an algorithm called AnyOut for computing outliers from data stream anytime. The AnyOut algorithm builds a precise tree topology- ClusTree to identify outliers at any time whether the data are constant or varying. ClusTree is a special feature of the model, it plays a part for creating the clusters.

A clustering-based approach using k-means was proposed by Elahi et al. (2008), it detects outliers by splitting data streams into chunks. Although the model doesn't perform good for grouped outliers. They experimentally presented following: comparison with some existing approach (Angiulli and Fassetti, 2007; Pokrajac et al. 2007) the model has improved performance for investigating outliers from data streams. The authors suggested that combining distance-based methods with their clustering model will yield better results. However, the model merely discovers the outliers but doesn't assign any outlier scores. MacQueen (1967) presented a pipeline to investigate outliers in varying data streams by utilizing similar approach as k-means. The model assigns weights for each feature based on their significance. The weighted features are significant, during algorithm processing they restrain noise effect. Comparing the algorithm with LOF (Breunig et al. 2000), it showed better detection rate including low time dissipation and low false positive rates. However, the algorithm doesn't define the degree of outlireness, therefore it might be a good future study to extend the pipeline and make it scalable over different data types. Later in another study, Morady et al. (2013) tried to implement cluster-based algorithm for big data, applying k-means algorithm to build an advanced pipeline, it was deemed succesful.

Bhosale (2014) combined both partitioning and distance-based approach to build an unsupervised model for data streams. They used partitioning clustering scheme (Ng and Han, 1994) that provide weights to the clusters according to their adaptivity and relevance by utilizing weighted k-means clustering. The concept of the model can evolve and adapt incrementally. The authors mentioned that it has higher OD rates than Elahi et. Al. (2008), and they suggested to include both categorical and mixed data as part of a future study. Another interesting method proposed by Moshtaghi et al. (2014), showed a clustering algorithm that can identify outlier beyond the cluster boundary. To observe the primary change is data stream distribution, the model continuously updates mean and covariance matrices. In another study by Moshtaghi et al. (2015), they proposed another framework on top of their previous one (Moshtaghi et al. 2014). The authors applied elliptical fuzzy logic to model the streaming data, to identify outlier - fuzzy parameters are updated by same style similar as Moshtaghi et al. (2014). For evolving dataset, Salehi et al. (2014) implemented an architecture based on ensemble learning. Ensemble methods create several clustering models instead of modeling the data streams and updating it from time to time. Evaluating all the clustering models, few are selected to measure the degree of outlierness for each data points. An efficient algorithm, based on clustering technique, is proposed by Chenaghlou et al. (2017). It showed improved memory usages and lower run time by presenting the concept of an active cluster. For any given data, they are divided into chunks where active clusters are analyzed in each chuck of data, the underlying data distribution also gets revised. Rizk et al. (2015) implemented an algorithm that investigates outliers in both small and large clusters. In another study, Chenaghlou et al. (2017) modified the method to perform detection in real time by Chenaghlou et al. (2018). Additionally, the model can detect cluster evolution sequentially. An effective algorithm, a cluster text OD algorithm, is proposed by Yin and Wang, (2016). If the chance of recognizing a cluster is low, it's highly probable to be an outlier. The model presents a technique (GSDPMM: Gibbs Sampling of Dirichlet Process Multinomial Mixture) to find if a document that held in a cluster is an outlier. Relating GSDPMM with incremental clustering can be a worthy research direction, as GSDPMM has a potential in incremental clustering. Later, Sehwag et al. (2021) proposed a unique framework called Self Supervised Detection (SSD) based on unlabeled distributions. They experimentally showed that their method, when it comes to unlabeled data, outperforms some of the traditional OD algorithms and even performs better than supervised detectors.

### 2.3.1. Advantages of Clustering-Based Methods

Clustering-based methods are unsupervised, therefore if underlying distribution knowledge is not necessary, then these models are a suitable choice. After the models learn about the clusters, they can test additional datapoints for detecting outliers. Again, unsupervised nature is suitable for incremental model as underlying distributions aren't required. They are robust algorithms and can manage versatile data types. For example, the hierarchical clustering methods for OD are good choice for different data types, they produce nested multiple partitions which is helpful for users to select partitions belongs to a certain level.

### 2.3.2. Disadvantages of Clustering Based Methods

A major drawback of clustering-based algorithm is that the outliers aren't assigned a score but binary labeling, where score represents degree of outlierness for a sample. Scoring is necessary because it helps to back track model actions, therefore the actions of a model become final and cannot be undone. Declaring the best number of clusters initially is a difficult job and most of the clustering algorithm often face difficulties with it. Also, if the cluster shape is arbitrary, the algorithms face problems understanding exact clusters from a given dataset. Therefore, to perform well, the shapes of several clusters need to define initially although it is a daunting task to provide the shape and distribution of multiple clusters. Partitioning-based methods are very sensitive to initialization of parameters like density-based methods. Nevertheless, they are inadequate to describe clusters and in most cases are not suitable for very large dimensional datasets. Additionally hierarchical based clustering methods showed expensive simulations in methods proposed by Karypis and Han (1999) and Zahn (1971) which makes them a poor choice for large datasets.

### 2.3.3. Research Gaps and Suggestions

It is important to note that, when designing any cluster-based models several questions need to be answered. An object defined as outlier, is it belongs to a cluster, or it located outside of the cluster boundary? If the distance between the object and the cluster centroid is distance, can it be labeled as outlier? If an object fits in a sparse or insignificant cluster, how the labeling can be performed within the cluster? Although clustering-based models have several drawbacks, they are good choice for most cases. Data stream is an interesting area for many researchers to apply cluster-based algorithms. For hierarchical and partitioning based clustering methods, speeding up the calculation process for large dataset and reducing CUP usage could be a suitable research direction. Detecting outliers from lower density populations or within a low density cluster can make the algorithms robust.

### 2.4. Distance-Based Methods

Distance-based OD methods are popular in many application domains, the foundational technique behind this method is nearest neighbor model. A straightforward example of this method would be to apply KNN to a dataset, and based on distance of a data point, it's either reported as an outlier or non-outlier. By closely relating to density-based assumptions, distance-based methods have underlying assumptions that outlier points KNN distances are large compared to normal data points. In contrast, with clustering-based approach, they are more granular in their analytical procedure. Therefore, these models are more effective in separating strong and weak outliers from malicious datasets. Again, referring to Figure 1, it is evident that clustering-based methods face difficulties detecting outliers in noisy data. According to the definition of clustering-based outlier definition, outlier point "A" and nearest centroid of a cluster will be similar for both Figures 1(a) and 1(b). In contrary, distance-based methods consider distances from point "A" and noisy data are handled accordingly in terms of distance estimation. However, cluster-based methods can be modified to address the issue of noisy samples, in that case, these two methods have the same organization as they are closely related. The distance-based algorithms provide scores to each data points incurring operational complexity proportional to $O(n^2)$. If binary labeling is expected as the outcome of the model, pruning techniques can be used to speed up the model substantially.

### 2.4.1. K-Nearest Neighbor Models

KNN is one of the fundamental algorithms for distance-based OD approaches. Initially, nearest neighbor methods detect global outliers and then assign them outlier scores. In KNN classification, it investigates distance information form a point to its neighbor, whether it's close or not. The fundamental idea is to utilize distance estimation to identify outliers. Knorr and Ng (1998) proposed a novel approach based on non-parametric technique that showed significant improvement over state-of-the-art OD algorithm at the time, especially for large dataset. Their approach differs from some of the previous method proposed by Yang, Latecki, and Pokrajac (2009) and Satman (2013) where a user doesn't know about the underlying distribution of the dataset. Their computational complexity is $O(kN^2)$, where N is the number of the datasets and k is the dimensionality. In Knorr and Ng (1998), nested loop and indexed based algorithm were applied to design OD models. Afterwards, Ramaswamy et al. (2000) proposed an improved technique that addressed the shortcomings of OD model by Knorr and Ng (1998), addressed computational cost, ranking method, and distance. They adopted the k*th* nearest neighbor that helps to ignore assigning distance parameter for the OD model. In another study, Knorr et al. (2000) expanded OD model proposed by Knorr and Ng, (1998), modified nearest neighbor estimation by applying X-tree, KD-tree, R-tree and indexing structure. For each example, the index structure is queried for nearest k points. Finally, top n number of outlier candidates are selected. However, the model falls apart when applied to large dataset of index structure.

Angiulli et al. (2006) proposed a technique that detects top-n number outliers from an unlabeled dataset. After that, the model predicts if a particular point is either an outlier or not. Top outliers get the highest weights, this is done by observing if a sample's calculated weight is higher than the top-n highest weights. Their approach incurs an $O(n^2)$ computational complexity. Later, Ghoting et al. (2013) developed an algorithm to address drawbacks of OD methods by Knorr and Ng, (1998) and Ramaswamy et al. (2000), where they tried to improve the run time for high dimensional feature space. They named the model Recursive Binning and Re-Projection (RBRP). In 2009, Zhang, Hutter, and Jin (2009) took a different path and projected an algorithm called Local Distance-based Outlier Factor (LDOF) that manages local outliers. Their study presented significant improvement compared to LOF (Breunig et al. 2000) in terms of range of neighbor size. This algorithm is similar in performance to KNN OD methods such as COF (Tang et al. 2002), however sensitivity on parameter value is insignificant. Later in 2013, a new model called Rank-Based Detection Algorithm (RBDA) is proposed by Huang et al. (2013) to rank neighbors. It understands the meaning and nature of high dimensional dataset by providing a feasible solution. The key assumption of the model is: objects will be similar and close to each other thereby sharing similar neighborhood if they are generated from the same apparatus. Instead of taking object distance information from neighbors, the model considers individual objects ranks which are close to the degree of proximity of the object. Another method, proposed by Bhattacharya et al. (2015), applies reverse nearest neighbor and nearest neighbor as an extended study of RBDA.

Dang et al. (2015) applied an OD algorithm using KNN in large traffic data in big cities. The model they proposed detects outliers by exploiting the information among neighborhoods that outliers are far from neighbors. This pipeline shows improved accuracy (95.5%), which is better than some statistical methods such as GMM (80.9%) and KDE (95%). Despite improved accuracy, is has trouble keeping a single distance-based measure. Wang et al. (2015) used a least spanning tree to increase searching mechanism of neighbors of KNN algorithm. In another paper, Radovanović et al. (2015) proposed a reverse nearest technique to manage high dimensional feature space. They presented the pipeline that can both manage low and high dimensional datasets. In terms of OD rates, this method works better than the original KNN method presented in Ramaswamy et al. (2000). Their method shows good performance on high dimensional datasets. In contrast to OD model proposed by Ramaswamy et al. (2000), Jinlong et al. (2015) modified a technique to get the neighborhood information using a natural neighbor concept. In another study, Ha et al. (2015) implemented a heuristic technique to achieve k value by employing random iterative sampling. Recent study on OD in local KDE is investigated by Tang and He (2017). Several types of neighborhood information were examined by them including k nearest, shared nearest, and reverse nearest neighbor. The KNN-based approaches are easy to implement despite of their sensitivity to parameter selection and less superior performance.

### 2.4.2. Pruning Techniques

Pruning technique is popular tool in ML models. A method, utilizes pruning technique method and randomization rule, based on nested loop, is presented by Bay et al. (2003). They modified the nested loop technique which was earlier known as quadratic $O(n^2)$ in performance and transformed into almost linear for most of the datasets. However, various assumptions in this pipeline resulted in poor performance. In another study, Angiulli and Fassetti (2007) presented a generic pipeline, where outliers are detected by pushing data in an index. While developing the algorithm, they focused on minimizing input and output cost as well as CPU cost, because these costs were a major challenge in previous research (Knorr et al. 1998, 2000; Ramaswamy et al. 2000), where they achieved both demands simultaneously. Ren et al. (2004) implemented a model to improvise the model proposed by Ramaswamy et al. (2000), they added pruning and labeling techniques to present a vertical distance-based OD algorithm. The method is implemented on both with and without pruning method while adopting P-tree. Applying P-tree technique to other Density-Based OD can be a good future work. Later, another technique is developed to improvise OD model proposed by Ren et al. (2004) for speeding up the detection process by Vu et al. (2009), where similar pruning techniques are applied.

### 2.4.3. Time Series Data

Time series continuous data naturally create problems such as uncertainty (Shukla, Kosta, and Chauhan, 2015), multidimensionality, notion of time and concept drift while applying them to an OD model. Usually, time series data are segmented by a time window, two popular time series window methods are: sliding window (Angiulli et al. 2010) where two sliding endpoints are used to mark a window, and landmark window where time points are identified to analyze *from-to* timeframes. A novel pipeline, proposed by Angiulli et al. (2010), utilizes distance-based approach where three different algorithms were developed for OD in time series data. They named the pipeline: STORM short for Stream Outlier Miner, exact STORM utilizes two modules: data structure and stream manager, where the later collects continuous data streams and the former one is applied by the stream manager. However, sorting cost of window is a shortcoming of the algorithm and colossal memory creates a burden as it cannot fit properly into memory. Later, Lai et al. (2021) performed OD time series benchmarked dataset and defined new context aware OD.

In another study, Yang et al. (2009) developed several methods: Extra-N, Exact-N, Abstract-C and Abstract-M to detect outliers based on neighborhood pattern information in the sliding window. This approach makes proper use of incremental OD by utilizing neighbor pattern in the sliding window of the dataset, which was not studied in earlier algorithms such as DBSCAN (Zhang, 2013). This algorithm shows improved performance, linear memory utilization per objects in a sliding window along with lower computational cost. Abstract-C applies a distance-based approach while Extra-N, Exact-N and Abstract-M utilize density-based cluster methods.

In another study, Angiulli and Fassetti (2007), several issues were discussed in event detection which were tackled by Kontaki et al. (2011), along with sliding window issues on time series data (Yang et al. 2009). Angiulli and Fassetti (2007) applied step function for processing the OD, whereas two algorithm parallelly utilizes the sliding window. Primary focus in Kontaki et al. (2011) was to make the method flexible, lower storage usages and enhance model efficiency. To support these ideas, three algorithms were proposed: COD, ACOD and MCOD short for Continuous, Advanced Continuous and Micro-cluster-based Advanced OD respectively. COD has two version which support multiple values of k and a fixed radius R, where k and R are the parameters for OD algorithm. On the other hand, both multiple radius and k values are supported by ACOD. MCOD needs less distance calculation done for OD by minimizing query range. COD, compared to STORM and Abstract-C algorithm, reduces the number of objects in each window and requires less memory space. Another method was developed to process large data volume proposed by Cao et al. (2014), it optimizes the range queries by not storing the objects in same window of same index structure. It is experimentally proven by the authors that MCOD is the most successful performing OD among COD, ACOD and MCOD.

### 2.4.4. Advantages of Distance-Based Methods

These methods don't rely on underlying distribution of data to detect outliers, thereby are straightforward algorithms. They also perform better compared to statistical-based methods and scale well for high dimensional dataset because of their robust architecture.

### 2.4.5. Disadvantages of Distance-Based Methods

Although distance-based methods perform better on high dimensional feature spaces than statistical-based methods, the increasing dimensions issue reduces their performance. This is because different objects has distinctive attribution in the given dataset, which make it difficult for the model to measure distance among such objects. Also, if KNN is applied for computing distance-based OD, then the model becomes computationally expensive and unscalable. For data streams, distance-based methods face difficulties in both data distribution in local neighborhood and investigation of KNN in the time series data.

### 2.4.6. Research Gaps and Suggestions

Distance-based algorithm are effective mathematical tools to sought anomalies in a dataset. One major challenge is to scaling for high dimensional dataset (Aggarwal et al. 2001). Very large feature spaces and object's random attributions force models to underperform. Not only increasing feature space reduces the ability of the model to describe by distance measures but also makes it difficult to comprehend the indexing approach to assigning neighbors. Additionally, multivariate data make the model less scalable when calculating distance measures. The models can be modified further by both improving execution time and memory usages. Another challenge is the quadratic complexity of the models, where researchers developed many techniques including pruning and randomization (Bay et al. 2003), compact data structure (Bhaduri et al. 2011; van Hieu et al. 2016). Distance-based methods are unable to detect local outliers, therefore often global information is calculated instead. To achieve desired scores from KNN algorithms, datasets need to be appropriate and properly processed. Selecting appropriate parameters including proper k value dictates performance of the model, and optimizing value of k and other parameters isn't easy always.

## 2.5. Ensemble Methods

In recent days, many domains such as healthcare and technology apply meta-algorithms for data mining problem such as classification or clustering to improve the solution. Such meta-algorithms create a series of multiple learning techniques: combinedly acts as a robust algorithm known as ensemble. Ensemble methods are mostly used in ML for their superior solutions compared to other traditional methods. These approaches are relatively new, and applied mostly on clustering and classification problems. The main idea behind this method is to train a dataset with multiple weak learners while each learning outcome gets improved by a subsequent learner, therefore reducing the loss function. This working architecture lets the model be independent of dataset localizations. Although, detecting outliers using ensemble is not straightforward, many algorithms are proposed in recent years: Bagging, Boosting, Bagged Outlier Representation Ensemble (BORE), Extreme Gradient Boosting OD (XGBOD) and Isolation Forest (Lazarevic and Kumar, 2005; Rayana and Akoglu, 2016; Micenková et al. 2015; Zhao and Hryniewicki, 2019; Liu et al. 2008). Bagging and Boosting algorithms solve classification problems, for sequential methods XGBOD is applied, for hybrid and parallel models BORE and Isolation Forest are applied.

One of the first ever ensemble method is known as Bagging, refined recently by Lazarevic and Kumar (2005), it shows improved performance over large dimensional dataset by utilizing feature bagging techniques. This technique splits and creates random subsets of features and combines the outcome of multiple detection algorithms applied separately onto the subsets of features. Each algorithm is randomly assigned a small subset of feature to provide an outlier score, these scores are labeled to all the datapoints. They experimentally showed that bagging has improved performance because it focuses on the outcome of multiple algorithms where each algorithm targets a small portion of a feature.

In another study, an ensemble method is presented for outliers' detection by Aggarwal et al. (2013) which was later discussed by many others (Kirner et al. 2017; Campos et al. 2018). Others proposed Bagging (Lazarevic and Kumar, 2005) and Boosting (Campos et al. 2018) from a classification context for ensemble analysis; also, alternative clustering (Müller et al. 2010) and multi view (Bickel and Scheffer, 2004) methods were proposed from a clustering context. Some critical questions were answered such as how to categorize, if ensemble methods are independent or sequential, and how to categorize, if ensemble methods are model- or data-centered? Ensemble algorithms are generally classified based on component independence. For instance, the components in boosting algorithms are not independent of each other because results in each stage depend on prior executions, whereas bagging is the opposite, which makes their components independent of each other. Also, if the methods are model centered, then the components of ensemble analysis are independent.

Later, several succeeding works have performed using ensembles for OD including Nguyen et al. (2010), Kriegel et al. (2011), and Schubert et al. (2012), which face various challenges. One of the issues is to score comparison provided by mixture models and various functions for outliers and combine them to get a general outlier score. In another study, Schubert et al. (2012), based on outlier scores, compared the outlier ranking by observing similarity events. Their approach is a greedy technique that achieved good performance through differentiating actions. In another study, Nguyen et al. (2010) addressed problems with high dimensional dataset and combined non-compatible OD method to form a unified approach. They implemented various scoring technique each time to determine the degree of outlierness of a sample instead of using same approach repeatedly. Because of their heterogenous approach they called their method Heterogenous Detector Ensemble (HeDES), which represents combination of functions and heterogeneity affair. The HeDES, in contrast to methods proposed by Lazarevic and Kumar (2005), assign score types and scores for different outliers. The method shows improvement on real-world dataset, however modification on the algorithm to handle large dimensional dataset can be a good research experiment.

Later in another study, Zimek et al. (2013) applied an arbitrary subsampling approach to calculate local density of nearest neighbors. When subsampling techniques are used on a dataset, usually training objects can be obtained without replacement, therefore they enhance OD performance. Also, subsampling technique with other OD can give good results as well. Zimek et al. (2014), later investigated an ensemble learning approach for OD, the pipeline brings a perturbation technique to account for different diversities in different outlier detectors as well as adopting a method that considers outlier rankings combinedly and distinctively.

As we suggested earlier, Pasillas-Díaz and Ratté (2016) did apply both feature bagging and subsampling technique together. Each technique is assigned to a different task: feature bagging extracts various information during each iteration, whereas subsampling technique scores different sets of data. However, getting variance of objects by using feature bagging was a drawback and the result depends on the size of the subsample. Except these shortcomings, the method has improved performance. Another method that dynamically combines the score values, an unsupervised framework, is proposed by Zhao and Hryniewicki (2019), they developed a way to combine and select outlier scores even if the ground truth is absent. Zhao et al. (2018) proposed a similar approach as Zhao and Hryniewicki (2019), and implemented four variations of it.

### 2.5.1. Advantages of Ensemble Methods

The ensemble analysis is better for investigating outliers because of their much better prediction models. Bagging and Boosting are two popular and efficient algorithms. They are robust and less dependent on a particular dataset in data mining processes. Ensemble methods are suitable for adopting high dimensional datasets, which used to be a burden for traditional OD algorithms.

### 2.5.2. Disadvantages of Ensemble Methods

Mathematically, ensemble analysis isn't that much robust as other data mining techniques, it is because they are not properly developed yet. This results in poor feature evaluation along with difficulties in selecting contextual meta-detectors. Various algorithms are combinedly working together and since the sample space is smaller, researchers face challenges managing real data in some cases using these methods.

### 2.5.3. Research Gaps and Suggestions

Although ensemble analysis has shown robust results, there are still issues that need to be fixed. They show good performance when streaming data has noise in it, because individual classifiers face difficulties when it comes to the quality of data and processing time. However, combinedly, those classifiers yield good outcome. Zimek et al. (2014) addressed multiple challenges along with data quality and processing time which has brought under consideration by developing models as Nguyen et al. (2010), Aggarwal and Sathe (2015), Liu et al. (2008), and Kriegel et al. (2011) to improve ensemble analysis for detecting outliers. Also, several research gaps have been addressed by Zimek et al. 2014, although ranking outliers from different detectors and diversifying principal proposals remains an open research challenge. Several techniques (Zimek et al. 2014; Rayana and Akoglu, 2016) don't require detector selection process, therefore these methods, in absence of detector selection process hardly help in speeding up identifying unknown outliers.

## 2.6. Learning-Based Methods

Learning-based methods are applied to different sub-discipline in ML. In this section we discuss four categories: Subspace, Active, Graph based and Deep Learning (DL).

### 2.6.1. Subspace Learning Models

OD models that have been discussed so far, usually identifies outliers from all the space and dimension, however outliers often represent different attributes in the local neighborhood on declining dimensional subspace. To address this issue, Zimek et al. (2013) presented that appropriate selection of a subset carries significant attribute information. On the contrary, residual attributes have less importance or sometime has no importance at all, and they delay the OD process. Subspace learning in OD is popular for high dimensional areas. The fundamental focus is to identify dissimilar dimension subsets and meaningful outliers form a given data. We can further categorize these studies into two subcategories: relevant subspace methods (Huang et al. 2013; Muller et al. 2008) and sparse subspace methods (Zhang et al.

2009; Dutta et al. 2016). The sparse subspace learning techniques project high dimensional datasets onto sparse and low dimensional subspace. The outliers are the ones located in sparse subspace because they are characterized as lower density. Projecting high dimensional space onto sparse subspace is time consuming therefore a big challenge. Aggarwal and Yu (2005) addressed this issue and proposed a method for effective subspace exploration, where an evolutionary algorithm gathers the subspaces. Here, initial population dictates the algorithms performance evaluation.

Later, Zhang et al. (2009) proposed a method that focuses on spares subspace technique's path. The method applies the idea of lattice to denote subspace relationship, sparse subspace is related to lower density co-efficient. Applying the idea of lattice makes the model perform poorly and complex in architecture. A new way to get sparse space is implemented by (Dutta et al. 2016), here sparse encoding is used to transform objects to multiple linear space. Relevant subspaces are used by outlier detectors to find local information as they are essential features in this case. A relevant subspace method is proposed by Huang et al. (2013) called Subspace OD (SOD). The method examines correlation of every object with its shared nearest neighbor; instead of taking distance from objects to its neighbors, the model considers ranks of each object that are close to the proximity of the object. Here, primarily the variance of the features is focused by SOD. Another method, in contrary to SOD, signifies the relationship between features is proposed by Müller et al. (2011).

In another but similar study, Kriegel et al. (2009) presented OD method that achieve relevant subspace where distances are computed by Mahalanobis technique through gamma distribution. Principal component analysis is used in this context. In contrast to Müller et al. (2011), the key difference is the requirement of large local dataset to recognize the abnormality trend. This impacts the scalability and flexibility of the method in a gradual manner. To tackle flexibility problem, a similar method is proposed by Keller et al. (2012) that identifies subspaces and ranks the outliers. The Monte Carlo method, a sampling technique, is implemented called High Contrast Subspace (HiCS), where LOF scores are combined based on HiCS values. In another study by van Stein et al. (2016), after achieving HiCS instead of using LOF scores, LoOP scores are used to calculate the degree of outlierness.

Nevertheless, the subspace learning methods are highly efficient for OD in several cases, they are computationally expensive. Searching for subspaces in high dimensional space is a daunting task which makes the pipeline more complex.

### 2.6.2. Active Learning Models

Active learning methods are semi-supervised learners through input sources or by interacting with users to get the desired outputs (Das et al. 2016). For instance, for large dataset that require labeling, doing so manually is an exhaustive process. Since the method is querying the user iteratively, this supervised approach is called active learning. When an active learning algorithm is trained, it can find smaller portions of the dataset that contain the labels. This helps the algorithm to re-train and boost for improvements. Also, by querying labels for instances from the user iteratively, it provides better suggestions. Recently, researchers are focusing on this approach for OD in different domains (Zhang et al. 2009; Dutta et al. 2016; Yiyong et al. 2007; Muller et al. 2008). Aggarwal and Yu (2005) applied active learning to unveil the reason for flagging the outliers and the reason behind high computational demand for estimating density for OD methods. The sampling process that was applied is called ensemble active learning. Later, Görnitz et al. (2014) applied an active learning method for OD, they alternatively repeated the learning process and updated the model to improve prediction results. After training on improved and unlabeled examples, the active learning method is applied.

In another study, input from a human analyst is provided to get better result using active learning (Dutta et al. 2016; Yiyong et al. 2007). Although they selected good portion of instances for the querying process, they didn't provide any explanation or clear insight or interpretation for the model design procedure. However, later they attempted to address the issues, a modified active learning approach is proposed by by Das et al. (2019). They called the method Glocalized Anomaly Detection (GLAD). Their primary focus is to adopt ensemble outlier detectors so that they can solve active learning problems. The end users have the control to global outlier detector, GLOD attains the local weights of data instance by learning automatically. Here, label feedback helps to implement this process. Also, proper tuning of ensemble detectors helps to identify maximum number of accurate outliers. This pipeline is also known as human-in-the-loop, where label feedback is achieved by a human analyst in each iteration round. In another study, Zha et al. (2020) proposed a deep reinforcement learning-based OD algorithm, detects outliers by achieving balance between long- and short-term rewarding processes.

Even though active learning serves a great purpose in OD community, there are still scope for improvement. Receiving inputs from human analyst is a daunting task, an AI assurance method is required to minimize the effect of false positive labeling while designing the model. Active learning methods are better at identifying outliers, however more interpretation techniques should be adopted to explain the results.

### 2.6.3. Graph Based Learning Models

Graphs are known as data structure that can adapt various algorithm, especially neural network, to perform learning task such as clustering, classification, and regression. Applications of graph data are getting popular for OD in various sectors. Initially, these algorithms transform each vector node into a real vector. Then the outcome is a vector representation of each node where information gets preserved in the graph. After achieving a real vector, one can apply it to a neural network.

Many algorithms have proposed especially OD in graph data, a broad review of graph-based OD approaches are presented by Akoglu et al. (2014) and Ma et al. (2021). The authors have presented state of the art techniques and several research challenges. They also discussed the importance of using graph-based OD, where graph-based approach shows the interdependency state of the data, robust and insightful distribution. A very first graph-based detection framework is called "*Outrank*" proposed by Moonesinghe and Tan (2008). They established entirely undirected graphs using the original dataset and a technique is applied to the predefined graph called Markov random walk. Markov random walk stationary distribution values are used to score all samples. Later, a novel approach is presented by Wang et al. (2018), where objects local information together with combined representation of the graph is adopted. They addressed the issue of false positive rates in OD, where graph-based method ignores local information of an object around each node. Therefore, local information of each object's surrounding of each node is collected that helps to construct the graph. Afterword, outlier scores are provided by randomly "walking through" the graph. This method adopts multiple neighborhood graphs where outlier scores are generated by walking through predefined graph. The authors concludes that their model showed good improvement. The Graph-based OD methods are relatively new and promising technique which has great potentials for OD in many domains.

### 2.6.4. Deep Learning Models

Deep Learning (DL) methods are a member of the ML family that are mainly applied for representation and patterns' learning by incorporating Artificial Neural Networks (ANN). Application of DL can be supervised, unsupervised or semi-supervised. These methods are getting popular because of their high accuracy on detecting outliers in critical infrastructure, healthcare, and defense (amongst many other domains). A survey in DL presented by Chalapathy and Chawla (2019) reviewed multiple DL-based OD technique and their evaluation. These models are effective for large dimensional dataset and can understand hierarchical information on features. Additionally, they are better for separating the boundary conditions between normal and abnormal behavior in time series dataset. Supervised DL models explore outliers by training and classifying the relationship between features and labels. For example, supervised models such as multiclass classifier is used to detect fraudulent transaction in healthcare (Chalapathy and Chawla, 2019). Although, supervised models provide great results, but unsupervised and semi-supervised models are mostly practiced. This is because, supervised models require labeling for each sample, so it's a daunting task to label each sample. Therefore, unsupervised and semi-supervised models are a better selection in real-world application with big datasets.

Semi-supervised DL methods for OD is the most appealing approach, given it provides flexibility regarding labeling requirements. The models use normal instances as references to identify outliers. Deep

Page 24 of 38

autoencoder, a semi-supervised deep neural learning model that can be applied to a dataset to find outliers. If enough training sets with normal events can be provided, the autoencoder can understand the inter-dependency of features. It generates a *reconstruction error* for all input features by encoding and decoding them, where the abnormal instances have higher reconstruction error.

Unsupervised DL OD techniques focus on essential features to find outliers form dataset. They label the dataset which is initially not labeled. The autoencoder is a popular unsupervised DL OD technique (Chen et al. 2017). In recent research (Zhou and Paffenroth, 2017; Chalapathy et al. 2017), unsupervised DL OD algorithm shows great effectiveness. Unsupervised models can be divided into two subcategories such as model architecture embracing hybrid models (Erfani et al. 2016) and autoencoders (Andrews et al. 2016). The autoencoder related models measures the degree of outlierness by observing reconstruction error of each feature space through adopting the value of residual vector. Hendrycks et al. (2018) implemented an approach for improving OD technique called Outlier Exposure. They identified a classification model by performing iteration to understand the heuristics, it helps to distinguish between distributed samples and outliers.

A universal framework that utilizes DL technique to log online OD and analysis called Deeplog presented by Du et al. (2017). To model the system log, Deeplog applies Long Short-Term Memory architecture. The algorithm learns and encodes the whole logging process. In contrary to other method where outliers are detected in each session, Deeplog learns outliers for every log entered. In high performance computing system, Borghesi et al. (2018) developed OD technique using autoencoder (Neural Network). A set of autoencoders are trained with the supercomputer nodes to learn the normal behavior, afterwards those autoencoders can identify abnormal behaviors.

Based on training mechanism, deep leaning OD methods can engage either One Class Neural Network or Deep Hybrid Models (Chalapathy and Chawla, 2019). Adopting deep neural networks, Deep Hybrid Models mainly emphasize on extracting feature from the autoencoder and after learning the hidden representation from the autoencoders. Most OD algorithms use them as inputs such as One Class SVM. Because of the shortage of labeled datasets for OD, hybrid approaches have notable limitations despite their performance maximization for OD. Therefore, features that are rich and differentiable are applicable for Deep Hybrid Models. To address and solve this problem, Ruff et al. (2018) introduced Deep One class Classification and Chalapathy et al. (2018) introduced One Class Neural Network.

### 2.6.5. Advantages of Learning-Based Methods

In graph-based approach, inter dependency of datapoints gets revealed by exhibiting an intuitive representation for OD. DL methods however are good for investigating the hierarchical discrimination between features in each dataset. Also, they have improved performance on large dimensional time series data. For time series data, they have effective ways to set boundaries between normal and outlier data.

### 2.6.6. Disadvantages of Learning-Based Methods

Learning-based model, especially subspace learning is computationally expensive. Generally, not all traditional DL methods are good on increasingly large amount of feature spaces, therefore detection of outliers could become more challenging.

### 2.6.7. Research Gaps and Suggestions

Not all methods in neural network can effectively differentiate the boundary between normal and outlier points, which is a vital task for data mining. Moreover, further research is required for Recurrent Neural Networks, Long Short-Term Memory, Deep Believe Network for OD. Kwon et al. (2019) and Chalapathy and Chawla (2019) are surveys on deep neural network OD that present further insights.

## 3. Tools for Outlier Detection

There are many of the shelf libraries and tools available to apply OD research and development. Among many tools, we include the most popular ones that are frequently used by the research community:

a) Scikit-learn (Python): Scikit learn is a well-known tool for AI research. This tool has some popular algorithms including Isolation Forest (Liu et al. 2008) and Local Outlier Factor (Breunig et al. 2000) etc.
b) Python Outlier Detection (PyOD) (Python): PyOD is another popular tool for OD in multivariate data. This library is widely used in academic research and some commercial purposes, it includes ensemble methods and several DL techniques (Ramakrishnan et al. 2019; Kalaycı and Ercan, 2018).
c) ELKI (Java): ELKI, stands for Environment for Developing KDD-Applications Supported by Index-Structures, is a Java based open-source platform for developing KDD applications and other data mining OD algorithms. The source code is written in Java, it provides benchmarking and simple fairness assessment test for the algorithms (Achtert et al. 2010).
d) Python Streaming Anomaly Detection (PySAD) (Python): PySAD is an open-source python-based library for streaming data to identify outliers. It contains a collection of algorithms including more than 15 online detector algorithm and two PyOD detectors setting for data (Yilmaz and Kozat, 2020).
e) Scalable Unsupervised OD (SUOD) (Python): SUOD works on top of PyOD, it's an unsupervised learning OD acceleration framework for large scale dataset training and predictions (Zhao et al. 2020).
f) Rapid Miner (Java): Rapid miner (Kalaycı and Ercan, 2018) is a Java based OD extension. It adopts unsupervised approach including COF (Tang et al. 2002), LOF (Breunig et al. 2000), LOCI (Papadimitriou et al. 2003), LoOP (Kriegel et al. 2009).
g) MATLAB: MATLAB is a user-friendly commercial software that supports many OD algorithms.
h) Time-series Outlier Detection System (TODS) (Python): It's a python based full-stack environment for detecting outliers in multivariate data streams (Lai et al. 2020).
i) Skyline (Python): Skyline detects anomalies in near real-time.
j) Telemanom (Python): Telemanom adopts Long Short-Term Memory architecture for multivariate time series data to detect outliers.
k) DeepADoTS (Python): A collection of DL benchmarking pipelines for OD for time series data.
l) Numerical Anomaly Benchmark (NAB) (Python): For real-time and streaming data, NAB is used to evaluate multiple algorithms for benchmarking purpose.
m) Datastream.io (Python): Datastream.io is an open-source tool for detecting outliers in real time data.

## 4. Datasets for Outlier Detection

In data mining problems, two types of data are used to train any OD models including real data and synthetic data. Real data are expensive to generate and distribute because of their security and commercial aspect. In this chapter, we enlist multiple real datasets to begin modeling OD problems. Some of the most popular OD datasets are as following:

a) University of California Irvine (UCI) Repository: The UCI repository (archive.ics.uci.edu/ml) provides more than hundreds of datasets where many researchers use these datasets for evaluating their algorithm. However, this server mostly contains dataset for classification algorithms.
b) ELKI Dataset: (elki-project.github.io/datasets/outlier): ELKI has numerous available dataset that can be used for different type of OD algorithm and for assessing model parameters.

c) Outlier Detection Datasets (ODDS) (odds.cs.stonybrook.edu/#table1): ODDS contains various type of dataset and it's a good source for training-testing OD algorithms. Some of the popular datasets from this server are time series multivariate and univariate datasets, high dimensional data, and time series graph data.
d) Anomaly Detection Meta-Analysis Benchmarks (ir.library.oregonstate.edu/concern/datasets/47429f155): Oregon State University has enriched datasets for evaluating various OD algorithm.
e) Harvard Database: (dataverse.harvard.edu/dataset): This server contains datasets that can be used for benchmarking unsupervised algorithm. It also contains several datasets for supervised OD models.
f) Skoltech Anomaly Benchmark (SKAB) (github.com/waico/skab): This repository contains approximately 34 datasets, although authorities plan to add more than 300 datasets in near future for collective anomalies and point anomalies.

All the above-mentioned sources provide many collective datasets to begin with OD studies. However, most of the real-world datasets are not available publicly, because of security and privacy concerns. For instance, data from critical infrastructure such as Electricity Transmission, Water Distribution, and Healthcare aren't available publicly. Therefore, synthetic data are an alternative and next best option for creating specific domain related models. For example, BATADAL (batadal.net/data.html) presents a synthetic data by creating virtual Supervisory Control and Data Acquisition System (SCADA) on top of a water distribution system network (Daneels, 1999). Since most real SCADA data aren't publicly available, this synthetic dataset is a good choice for researchers. In data mining problems, various evaluation techniques are implemented for the OD algorithms to measure "goodness". These evaluation techniques focus on OD rates and run times of the algorithm. Mostly adopted evaluation measurements are Precision, R-Precision, Area Under the Curve (AUC), Average Precision, Receiver Operating Characteristics (ROC), Correlation Coefficient, and Rank Power (RP) (Domingues et al. 2017).

## 5. AI Assurance and Outlier Detection

In this chapter, we discuss several working algorithms for OD in data mining problems for AI assurance. According to (Batarseh et al. 2021), AI assurance can be defined as:

"A process that is applied at all stages of the AI engineering lifecycle ensuring that any intelligent system is producing outcomes that are valid, verified, data-driven, trustworthy and explainable to a layman, ethical in the context of its deployment, unbiased in its learning, and fair to its users."

Additionally, in their review paper, the authors added ten metric scoring schemes to present a systematic comparison among existing AI assurance approaches. To verify an AI model, six assurance goals need to be verified for an AI system: fairness, trustworthy, ethics, safety, security and explainability. The authors address the complexity of recent AI algorithms and the necessity of investigating algorithm variance, bias, clarity, and awareness to measure these AI assurance goals. In general, AI assurance goals can be achieved by either model specific or model agnostic approach. Model specific approaches target specific AI algorithms for quantifying or validating assurance goals, whereas model agnostic approaches are generic and have universal frameworks that can verify all AI algorithms for assurance goals. Despite the challenges, AI assurance is necessary, OD is at the heart of assurance as it improves the overall quality of the data.

Data quality needs to be assured as well. If the underlying data is invalid, then AI algorithms will have undesirable outcomes. OD algorithms measure two important aspects of data assurance: safety and security. This is because, analyzing a dataset for outlier not only means investigating abnormal samples but also represents faults or intrusions in the system by adversaries. For instance, ANN autoencoders detect outliers using a reconstruction error, where the errors are generated during encoding and decoding process of a dataset. Higher reconstruction errors are an indication of an object being an outlier or an attack on the system. Therefore, reconstruction errors, in this context can be considered as safety and security measure. Other data assurance goals can be achieved depending on the context of application domain and AI algorithm used.

Assurance goals, especially fairness and ethics can be achieved by removing bias in the dataset, however big data generated by real-world source almost always have bias (Verma et al. 2021). Some of the most common data biases are activity bias, selection bias, bias due to system drift, omitted variable bias, and societal bias. For identifying the reason behind any bias, one should investigate how the data are generated. Most common practice of data bias identification is to perform Exploratory Data Analysis (EDA) (Tukey, 2020). In recent study, Amini et al. (2019) presented a debiasing technique during post processing after training with AI algorithm. Their method adopts DL-based model to understand the latent data distribution during training stage in an unsupervised manner, thereby making the approach robust for debiasing. In another study, Bolukbasi et al. (2016) showed a debiasing technique to mitigate gender bias. For model assurance, in recent study, Shekhar et al. (2020) applied a novel framework based on deep-autoencoder for fairness called Fairness-aware OD (FairOD). They focused on formalizing the definition of fair OD algorithm with desirable properties. Data bias can yield unfair, unethical, and untrustworthy decisions by AI algorithms, therefore bias needs to be identified before training the AI model. Data bias can be also detected using OD algorithms.

## 6. Conclusions

This chapter reviews the state-of-the-art in approaches for outlier analysis. We group OD methods into several categories: Distance, Statistical, Density, Clustering, Learning and Ensemble-based methods. For each category we present relevant algorithms, their significant importance and drawbacks.

For distance-based methods, especially ones that use KNN based models, are sensitive to the parameter selection process including the value of k. Therefore, an appropriate k parameter selection is important for the models that rank neighbors for OD. Clustering-based methods generally are not explicitly suitable as they were not designed to facilitate OD, however ensemble methods that combine results from a collection of dissimilar detectors provide much improved outcomes. Ensemble methods have lower execution time but high-quality OD results. Regarding model evaluation, effectively assessing an OD algorithm is still an open research challenge. Also, in many cases, it's a daunting task to evaluate a model when a ground truth is absent and outliers aren't that frequent. Deep neural network-based OD models are gradually becoming popular because of their effective measures and quality results. ANN based autoencoders can detect outliers even if sensor network data are compromised and concealed by an adversarial attack. DL-based models are advanced and difficult to design nonetheless. Moreover, enough investigations are required to unlock the full potential of DL-based models for detecting outliers in real-world applications. Lastly, an important notion to note, OD models need to be assured, because AI algorithms ought to be safe and secure from unwanted outliers.